\documentclass[journal]{IEEEtran}

\usepackage{amsmath}
\usepackage{amssymb}
\usepackage{amsfonts}
\usepackage{bm}

\usepackage{graphicx}
\usepackage{epstopdf}
\usepackage{epsfig}
\usepackage{float}
\usepackage{tikz}
\usetikzlibrary{positioning, fit, calc, shapes.geometric, arrows.meta}

\usepackage[caption=false,font=footnotesize]{subfig}

\usepackage{booktabs}
\usepackage{multirow}

\usepackage{times}
\usepackage{indentfirst}
\usepackage{color}
\usepackage{url}

\usepackage{cite}
\usepackage{enumerate}

\usepackage{algorithm}
\usepackage{algorithmicx,algpseudocode}

\newtheorem{mydef}{Definition}

\usepackage[
    colorlinks=false,
    pdfborder={0 0 1},
    citebordercolor={0 1 0},
    linkbordercolor={1 0 0},
    urlbordercolor={0 0 1}
]{hyperref}

\begin{document}

\title{Less is More: Non-uniform Road Segments are Efficient for Bus Arrival Prediction}

\author{Zhen Huang*\thanks{* These authors contributed equally to this work.}, Jiaxin Deng*, Jiayu Xu, Junbiao Pang$\dagger$\thanks{$\dagger$ Corresponding author.}, Haitao Yu
\thanks{
\IEEEcompsocthanksitem Z. Huang, J. Deng, J. Pang and J. Xu are with the Faculty of Information Technology, Beijing University of Technology, Beijing 100124, China (e-mail: \mbox{junbiao\_pang@bjut.edu.cn}).
\IEEEcompsocthanksitem H. Yu is with the Beijing Intelligent Transportation Development Center,
Beijing 100161, China (email: yuhaitao@jtw.beijing.gov.cn).
}
}

\maketitle

\begin{abstract}
In bus arrival time prediction, the process of organizing road infrastructure network data into homogeneous entities is known as segmentation. Segmenting a road network is widely recognized as the first and most critical step in developing an arrival time prediction system, particularly for auto-regressive-based approaches. Traditional methods typically employ a uniform segmentation strategy, which fails to account for varying physical constraints along roads, such as road conditions, intersections, and points of interest, thereby limiting prediction efficiency.
In this paper, we propose a Reinforcement Learning (RL)-based approach to efficiently and adaptively learn non-uniform road segments for arrival time prediction. Our method decouples the prediction process into two stages: 1) Non-uniform road segments are extracted based on their impact scores using the proposed RL framework; and 2) A linear prediction model is applied to the selected segments to make predictions.
This method ensures optimal segment selection while maintaining computational efficiency, offering a significant improvement over traditional uniform approaches. Furthermore, our experimental results suggest that the linear approach can even achieve better performance than more complex methods.
Extensive experiments demonstrate the superiority of the proposed method, which not only enhances efficiency but also improves learning performance on large-scale benchmarks. The dataset and the code are publicly accessible at: \url{https://github.com/pangjunbiao/Less-is-More}.
\end{abstract}


\section{Introduction}

Accurate prediction of the Estimated Time of Arrival (ETA) for urban buses is a critical component of Intelligent Transportation Systems (ITS) and serves as a key enabler for improving public transportation service quality. Specifically, ETA prediction enhances passenger comfort levels~\cite{VARGHESE20252618}, optimizes scheduling efficiency~\cite{LU2025103461}, supports data-driven route planning~\cite{XIAO2024103787}, and improves operational and economic performance~\cite{VARGHESE20252618}. Consequently, ETA prediction plays a significant social role and has garnered sustained research interest~\cite{Ref3_Chen2013}.

Automatic Vehicle Location (AVL) technology has been significantly enhanced through the integration of Global Positioning System (GPS) in the bus industry. Consequently, large-scale GPS datasets have facilitated the rapid development of data-driven ETA prediction methods. For example, early studies predominantly employed statistical methods and machine learning methods~\cite{SERIN2022111403}, \textit{e.g.}, Kalman filtering~\cite{SHEN2025127622}; recently, Deep Learning (DL) techniques, including Recurrent Neural Networks (RNN), Long Short-Term Memory (LSTM)~\cite{PETERSEN2019426}\cite{Ref27_Pang2019}, and Transformer~\cite{Ref92}, have emerged as effective solutions for ETA prediction. Existing methodologies predominantly employ uniform interpolation of GPS observations to generate dense road segments~\cite{MA2019536}, or alternatively, utilize bus stations as proxies for segmentation~\cite{Ref51x}. A crucial factor lies in organizing road networks into road segments to effectively characterize the physical nature of spots (e.g., expressways, commercial districts, residential zones), which is often overlooked by previous works~\cite{Ref27_Pang2019}~\cite{Ref92}. 

\begin{figure}[t!]
\centering
\subfloat[Intersections without traffic lights]
{\includegraphics[width=0.45\columnwidth]{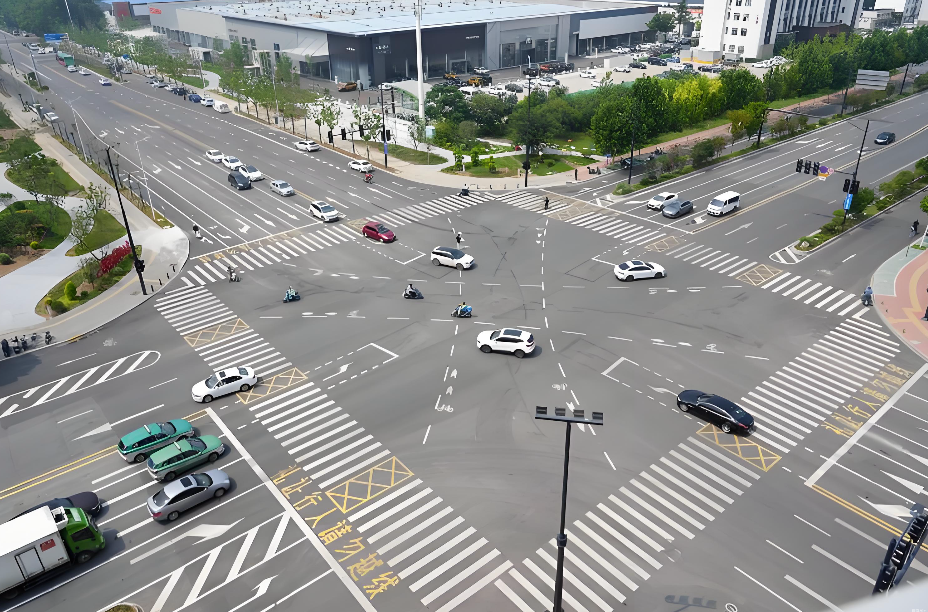}
}
\hfil
\subfloat[A middle school entrance during school hours]
{\includegraphics[width=0.45\columnwidth]{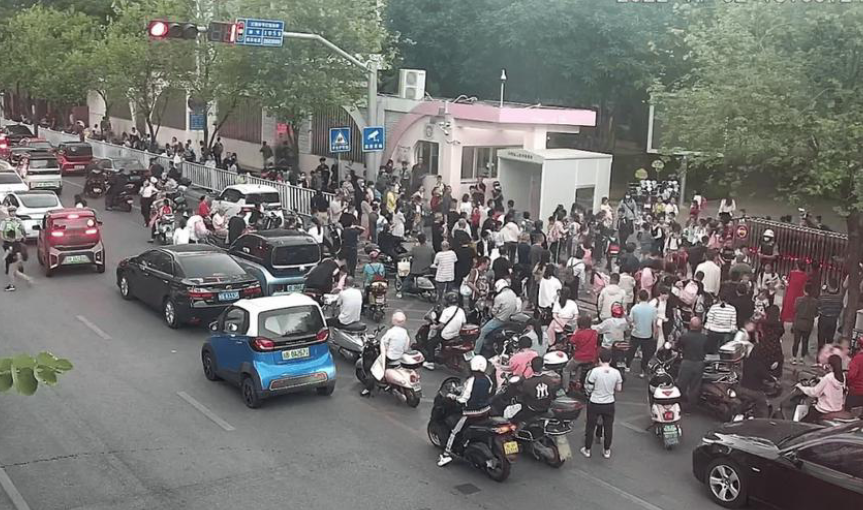}
}\\
\subfloat[Zebra crossing without traffic lights]
{\includegraphics[width=0.45\columnwidth]{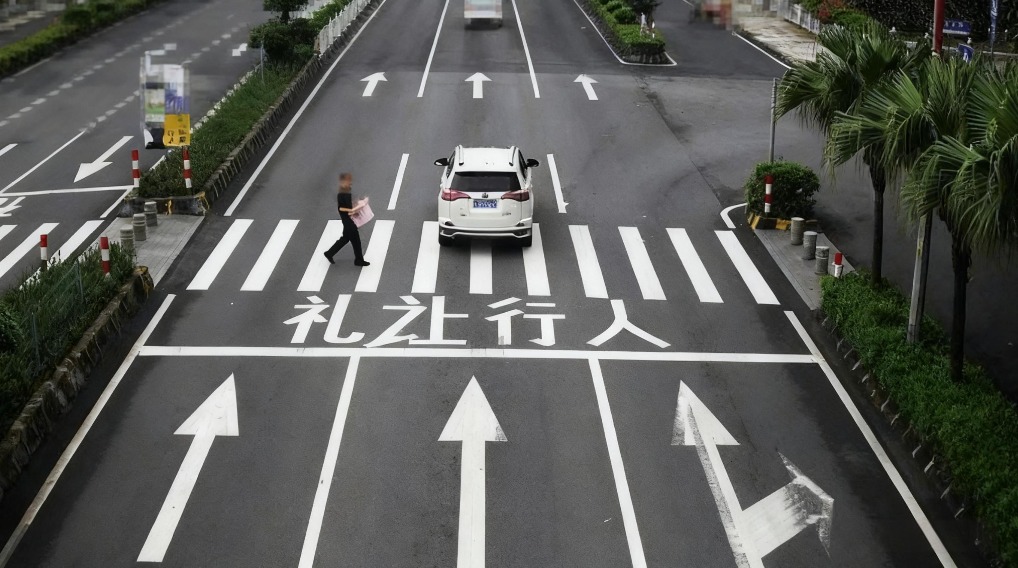}
}
\hfil
\subfloat[Main road with bus-only line]
{\includegraphics[width=0.45\columnwidth]{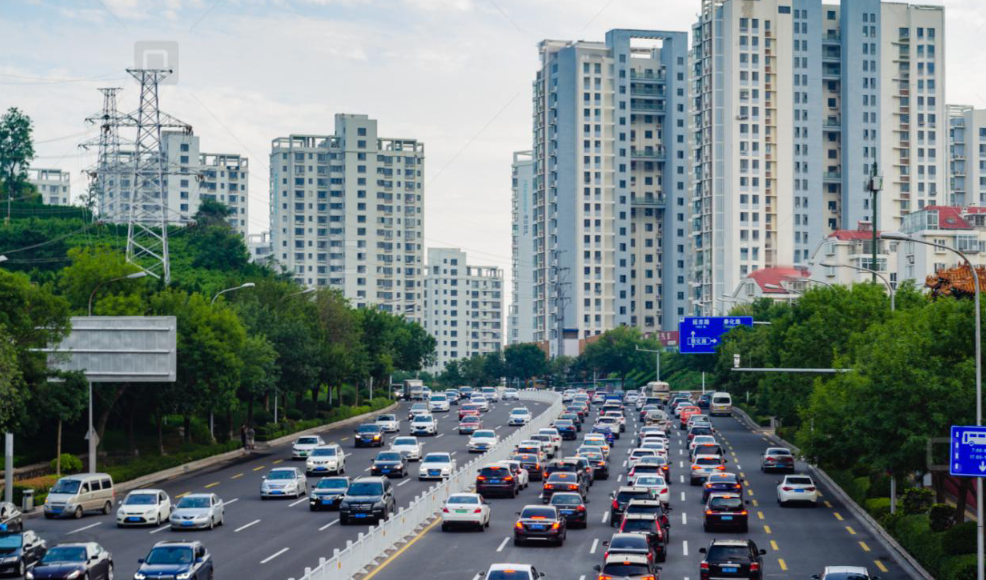}
}
\caption{Road segments significantly affect the bus ETA prediction. These spots have different temporal patterns to cause diverse predictability of bus ETA prediction.}
\label{fig:ETA_Restricted_Sections}
\end{figure}

\begin{figure}[t!]
    \centering
    \includegraphics[width=0.45\textwidth]{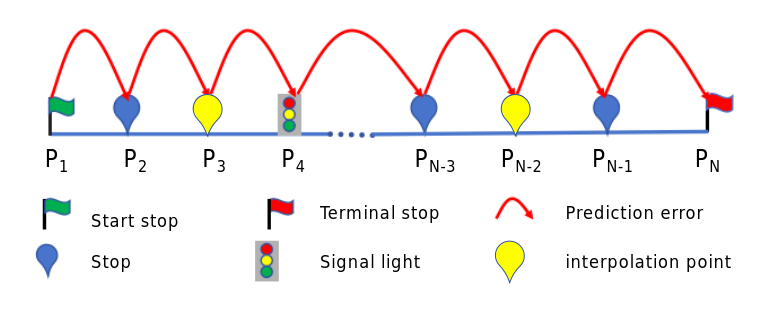}
    \caption{Prediction error is propagated into next prediction in auto-regression method.}
    \label{fig:prediction_error_propagation}
\end{figure}

The uniform interpolation strategy enhances the continuity representation of traffic states, still facing two critical limitations: 
\begin{itemize}
     \item \textbf{Uneven role of different road segments.} Uniform interpolation with fixed intervals fails to account for non-linear factors, such as intersections, resulting in redundant and less informative features. As Fig.~\ref{fig:ETA_Restricted_Sections} illustrates, these road segments at least are roughly classified into 4 groups: 1) Physically constrained spots (\textit{e.g.}, intersections with traffic lights, retail complexes); 2) Physically and temporally constrained spots (\textit{e.g.}, middle schools during school hours, and bus-only line); 3) Random factor constrained spots (\textit{e.g.}, zebra crossings without traffic lights). These spots are significantly more important for ETA prediction than road segments on the main roads.
     \item \textbf{Error accumulation is unavoidable in auto-regression-based ETA prediction.} Unnecessary interpolation points exacerbate the error accumulation issue within auto-regressive models, as discussed in prior work~\cite{Ref27_Pang2019}. As illustrated in Fig.~\ref{fig:prediction_error_propagation}, redundant road segments propagate prediction errors into subsequent predictions, significantly degrading the model's testing-time performance.
 \end{itemize}

Therefore, a natural question arises: \textit{given that not all sampling points contribute equally to the ETA prediction problem, would the model autonomously learn to focus on the informative ones?} This intuitive yet natural observation poses two major challenges: 1) The road segment selection and ETA prediction tasks are inherently interdependent; and 2) When trained jointly, gradient interference and conflicting optimization objectives may arise, leading to unstable training dynamics and degraded prediction accuracy.
Consequently, a progressive coordination mechanism between road segment selection and ETA prediction is essential for robust and reliable performance.

In this paper, we propose a Reinforcement Learning (RL)-based approach~\cite{LAI2024100164} to select informative road segments, progressively identifying those that contain representative traffic information. This method enables the ETA model to adaptively select informative features without compromising predictive accuracy, transforming static factors into a dynamic sequential decision-making process that captures essential traffic variations. Specifically, a novel reward function is proposed to jointly optimize the accuracy of ETA prediction and the sparsity of selected road segments. Furthermore, a progressive training scheme is introduced to integrate RL with a linear ETA model. The resulting ETA prediction model achieves a balance between efficiency and accuracy. The contributions of this work are as follows:
\begin{itemize}
    \item To the best of our knowledge, we are the first to demonstrate that non-uniform road segments are more effective for bus ETA prediction than uniform ones. Specifically, a Deep Progressive Reinforcement Learning (DPRL) framework is proposed to integrate a RL-driven Segment Selection Network (RSS-Net) and a linear ETA model, enabling collaborative optimization between dynamic road segment selection and ETA prediction.
    \item A reward function is introduced that jointly considers prediction error and feature sparsity, allowing the model to maintain or even improve prediction accuracy while significantly reducing the number of input features, thereby enhancing computational efficiency and real-time applicability for ETA prediction. 
    \item Even with a simple linear ETA prediction model, the proposed method matches or outperforms traditional approaches on public datasets. The improvement of the linear model demonstrates that not every road segment is useful for the transportation problem, shedding light on how to build road networks for traffic prediction~\cite{KHAIRY2025431} and generation~\cite{WANG202171} in ITS.
\end{itemize}

\section{Related Work} \label{sec:RW}

\textbf{Road network modeling for prediction.}
With the growing digitalization and interconnection of transportation systems, modeling traffic networks has become a paramount task for traffic prediction~\cite{KHAIRY2025431} and traffic generation~\cite{WANG202171}.
Existing solutions can be roughly divided into two paradigms: 1) the first incorporates physical characteristics of road nodes into the road network. For instance, the road network is constructed as a knowledge graph~\cite{Ref60}, which contributes to improving the accuracy of route prediction~\cite{JI2020578}; 2) the second integrates spatial relationships between road segments to characterize road networks. A typical example is that weighted undirected graph models for urban road networks are derived from traffic sensor data to support traffic state prediction~\cite{ref70}.

Despite significant progress in capturing spatiotemporal patterns within road networks, most existing research has predominantly focused on modeling road segment attributes~\cite{khan2025deep} or inter-segment relationships~\cite{zhao2019t}. In public transportation, the selection of informative road segments remains largely unexplored. Our paper addresses this gap by focusing on road segment selection rather than modeling road network attributes for bus ETA prediction.

\textbf{Bus ETA prediction.}
Traditional traffic prediction methods, \textit{e.g.}, linear regression models and historical average methods, often struggle to capture the non-linear variations and temporal dynamic characteristics of traffic flow.
For the bus ETA problem, these methods barely handle fluctuations in passenger demand and the influences of traffic signals effectively, thereby limiting their ability to achieve accurate predictions in complex traffic scenarios~\cite{ref72}~\cite{ref73}~\cite{ref74}.
 
In recent years, DL methods have gained attention for bus ETA prediction. These methods are roughly divided into two groups: 1) one notices the non-linear behavior of bus ETA prediction. For example, Ahsan and Siddique ~\cite{ref83} introduced a feedforward attention model that dynamically weights features like traffic density and road type. Roy et al. ~\cite{ref82} proposed a Deep Encoder Cross Network (DECN) that includes contextual factors beyond distance, \textit{e.g.}, traffic conditions and road characteristics, improving prediction accuracy over traditional baselines. 2) the other consider bus ETA on complex road network as prediction on graph. For instance, Petersen et al. ~\cite{PETERSEN2019426}introduced a spatiotemporal neural network that integrates GPS trajectory data with temporal features. 

Complex neural network structures often lead to high computational costs and poor real-time efficiency. In this paper, we employ a simple linear regression model combined with road segment selection, achieving comparable or even better prediction performance than state-of-the-art (SOTA) methods. The results highlight the importance of segment selection for traffic prediction tasks.

\textbf{Reinforcement Learning for Traffic Problem}
RL has garnered significant attention in traffic management, particularly for intelligent traffic signal control~\cite{farazi2021deep}. For example, early research primarily relied on Q-learning and limited state-space models in small-scale scenarios on hand-crafted rules ~\cite{ref86}. However, these methods fail to fully capture the complexity of traffic environments and ignore more relevant traffic information, leading to suboptimal performance in signal control~\cite{ref90}. Recently, RL combined with neural networks has been able to handle more complex and large-scale tasks. For instance, Li et al. proposed a deep Q-network to approximate the Q-function and optimize traffic signal control at intersections ~\cite{ref87}. Shabestary et al. employed deep neural networks to directly process sensory inputs for signal control strategies~\cite{ref88}. 

To our best knowledge, we firstly leverage RL to select road segments for ETA prediction by leveraging RL to achieve this goal, demonstrating superior adaptability and real-time performance in real-world traffic scenarios.

\section{Methodology} \label{sec:method}

\begin{mydef}[A journey of a bus] 
A journey of a bus \( \mathcal{T} \) consists of a series of continuous GPS points, where the time interval between any two adjacent GPS points does not exceed a fixed threshold \( \tau \), representing the sampling interval.
\( \mathcal{T} : \bm{p}_1 \rightarrow \bm{p}_2 \rightarrow \ldots \rightarrow \bm{p}_N \), and
\( 0 < \bm{p}_{i+1} - \bm{p}_i \leq \tau \) for \( 1 \le i \le N-1 \).
\end{mydef}

\begin{mydef}[Road segment]
A road segment $\mathbf{r}$ is an un-directed edge. Each road segment is associated with a unique ID $r_{id}$, a starting point $\mathbf{r}_{s}$, and an ending point $\mathbf{r}_{e}$. Each point consists of the longitude and latitude coordinates.
\end{mydef}

\textbf{Remark.} For the bus ETA prediction, a journey of bus is discretized into road segments, which is composed of both the interpolated points and bus stations, as illustrated in Fig.~\ref{fig:prediction_error_propagation}.

\begin{mydef}[The ETA prediction on sparse road segments]\label{def:problem}
Given a journey of a bus line $ \mathcal{T}  $ with the uniformed road segments $\mathcal{R}=[\mathbf{r}_0,\ldots,\mathbf{r}_i,\ldots,\mathbf{r}_N]$ where $N$ is the number of segments, the problem of the bus ETA prediction on sparse road segments is to simultaneously find a road segment selector $\phi_s(\cdot)$ and a prediction method $\phi_p(\cdot)$, as follows:
\begin{equation}\label{eqt:problem}
\begin{split}
\min_{\mathcal{\hat{R}},\mathbf{\theta}_s,\mathbf{\theta}_p} \   \ &\text{loss}(\phi_p(\hat{\mathcal{R}}),y;\mathbf{\theta}_p)\\
s.t.: \   \  
&\mathcal{\hat{R}} = \phi_s\left(\mathcal{R ,\mathbf{\theta}_s}\right), \\
& \|\mathcal{\hat{R}}\|_0=M, \\
& M\leq N, 
\end{split}
\end{equation}
where $\text{loss}(\phi_p(\hat{\mathcal{R}}),y;\mathbf{\theta}_p)$ is a predefined loss function in which $\mathbf{\theta}_p$ is the parameter, $y$ is the ground truth for each bus stop, $\mathcal{\hat{R}}$ is the selected road segments with function $\phi_s\left(\mathcal{R},\mathbf{\theta}_s\right)$ in which $\mathbf{\theta}_s$ is the parameter for the selection function, the notation $ \|\cdot\|_0$ is the number of segments in set $\mathcal{\hat{R}}$, and $M$ is the number of selected road segments, $N$ is the total number of road segments.
\end{mydef}

\begin{figure}[t!]
    \centering
    \includegraphics[width=0.5\textwidth]{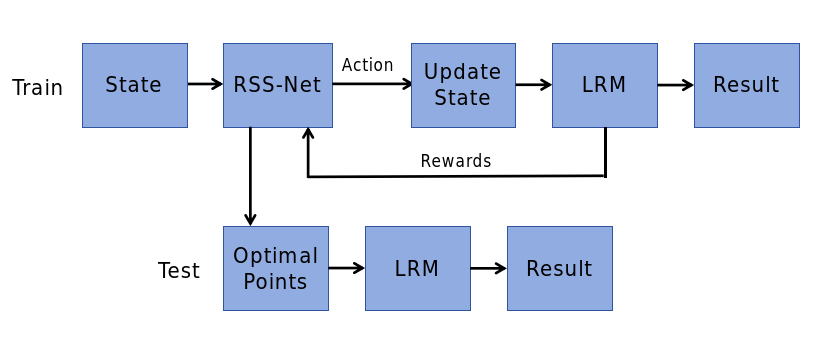}
    \caption{The architecture of the proposed method. It depicts the operational pipeline of the RSS-Net and LRM during the training and testing phases.}
    \label{fig:framework}
\end{figure}

Figure~\ref{fig:framework} depicts the overall framework of our proposed method. Specifically, our model comprises two key components: the Linear Regression Model (LRM) and the Road Segment Selection Network (RSS-Net). First, a route is linearly and uniformly discretized into road segments. RSS-Net leverages RL to find representative road segments. The selected segments are used to train a LRM. A reward signal is subsequently updated to refine RSS-Net. Thus, the ETA prediction model and RSS-Net iteratively interact and co-optimize, thereby enhancing the performance of bus ETA prediction.

\subsection{Linear ETA Prediction Model}

Given the $j$-th journey of a bus, we have time stamps ${t}^v_i$, (\( i = 0, 1, \dots, N, v=0,\ldots, V\)), where $t^v_i$ denotes the timestamp at the $i$-th road segment, $N$ and $V $ denote the road segment number and historical trajectories, respectively. Therefore, for a particular bus up to the $L$-th segments with the time stamps $\mathbf{t}^v_L = ( t^v_0, \ldots, t^v_i \dots, t^v_L )^{\top}$, our goal is to compute predictions \( \mathbf{t}^v_{L+H} \) for the times of arrival at locations \( L+H \), where \( H = 1, 2, \dots, N - L \). 

Let ${\sigma_{ij}}$ denotes the covariance between arrival times at the road segments $i$ and $j$, 
$\sigma_{ij} = \frac{1}{V} \sum_{v=1}^{V} (t_i^{v} - \mu_i)(t_j^{v} - {\mu}_j)$,  
where $\mu_i$ and $\mu_j$ are the mean arrival times at the segments $i$ and $j$, respectively, as follows:
\begin{equation}
\mu_i = \frac{1}{V} \sum_{v=1}^{V} t_i^{v}.
\label{eq:2}
\end{equation}

Let $ \mathbf{\mu}_L = (\mu_0, \mu_1, \dots, \mu_L)^{\top}$ and $\mathbf{\Sigma} \in \mathbb{R}^{L\times L} $ denote the mean vector and the covariance matrix of the first $L$ segments, respectively, where $\mathbf{\Sigma}_{ij} = \sigma_{ij}, i,j \le L$. 
Let $\mathbf{\sigma}_{L,L+H} = (\sigma_{0,L+H}, \sigma_{1,L+H}, \dots, \sigma_{L,L+H})^{\top}$ as the covariance vector between the segments $0, \dots, L$ and the target ones $L+H$. The predicted arrival time at segment $L+H$ is as follows:
\begin{equation}\label{eqt:LR}
\hat{t}_{L+H} = \mu_{L+H} + \bm{\sigma}_{L,L+H}^{\top}\mathbf{\Sigma}^{-1}(\mathbf{t}_L - \mathbf{\mu}_L).
\end{equation}
Eq.~\eqref{eqt:LR}  assumes that bus travel times follow a multivariate Gaussian distribution. Eq.~\eqref{eqt:LR} has a $O(L^2)$ time complexity. Note that the linear method can be replaced with any ETA prediction methods~\cite{Ref27_Pang2019}. In this paper, the simple linear model is used to verify the effectiveness of sparse road segments for ETA prediction. 

\subsection{Road Segment Selection Network}\label{rssn}

\begin{figure}[t!]
    \centering
    \includegraphics[width=0.45\textwidth]{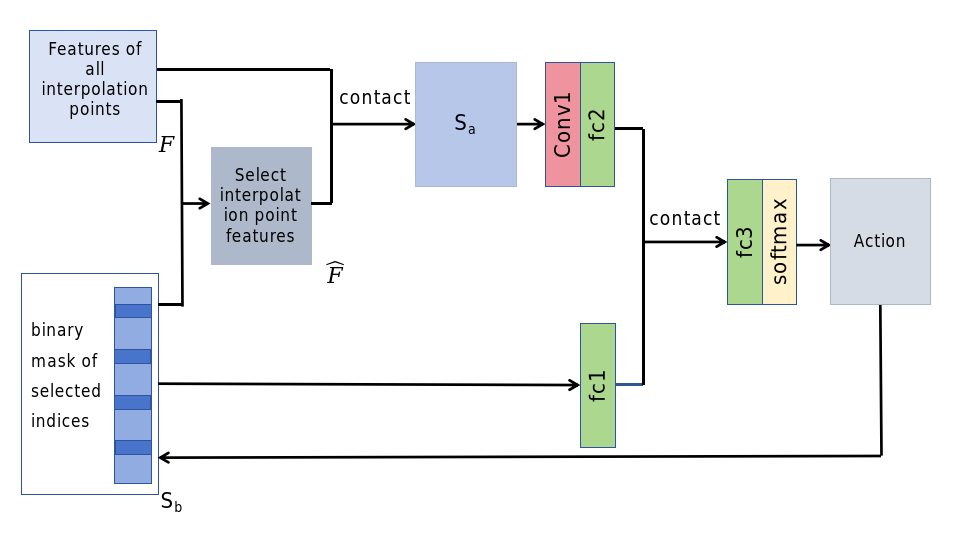}
    \caption{The neural network architecture of RSS-Net for adaptive segment selection.}
    \label{fig:ipsnet_structure}
\end{figure}

Fig.~\ref{fig:ipsnet_structure} illustrates the architecture of the proposed network. The input  consists of two components: 1) the feature coding of all road segments $\mathbf{F}$ ($\mathbf{F} \in \mathbb{R}^{N\times 8}$), and 2) a binary mask $\mathbf{S}_b$ ($\mathbf{S}_b \in \mathbb{R}^{N\times 1}$) indicating which segments are selected, \textit{i.e.}, $S_i=1$ means that the $i$-th segment is selected; otherwise, the $i$-th one is not selected. Concretely, the feature $\mathbf{S}_a$ is processed by a convolutional layer followed by a fully connected layer (fc1), while the mask $\mathbf{S}_b$ is independently processed by another fully connected layer (fc2). Two branches allow the network to model feature dependencies through two distinct pathways. The outputs of the two branches (fc1 and fc2) are concatenated and passed through a fully connected layer(fc3). Finally, the feature from fc3 is fed into a softmax function to obtain the probability of the actions. The details of network can be found in our publicly released code.

\textbf{Feature Coding of Road Segments:} 
For each segment, a feature $\mathbf{f}_i = \left[ t_0, t_i,  d_i, d_{i+1},r_i, \text{OHC} \right]^\top$ contains four categories of information as follows: 1) \textit{Temporal cues:} The departure time from the starting station $t_0$, and The travel time to reach the current segment $t_i$; 2) \textit{Spatial information:} The distance from the starting station to the current position $d_i$, and the distance from the next segment to the starting one $d_{k+1}$; 3) \textit{Static information:} The number of bus lines associated with the $i$-th segment, $r_k$; and 4) \textit{Position type:} One-Hot Coding (OHC) is used to distinguish the type of the current segment: \( [1, 0, 0]^\top \) for bus stop, \( [0, 1, 0]^\top \) for intersection, and \( [0, 0, 1]^\top \) for interpolated point.

\textbf{State:} The RL state $\mathbf{S}=(\mathbf{S}_a,\mathbf{S}_b)$ is defined as follows:
\begin{itemize}
   \item {\textit{Features of Road Segments}:} 
    $\mathbf{S}_a = [\mathbf{F}, \hat{\mathbf{F}}] \in \mathbb{R}^{(N+M) \times 8}$,  
    where $\mathbf{F} = [\mathbf{f}_{1}, \mathbf{f}_{2}, \ldots, \mathbf{f}_{N}] \in \mathbb{R}^{N \times 8}$ is feature for the all segments,  $\hat{\mathbf{F}} = [\hat{\mathbf{f}}_{1}, \hat{\mathbf{f}}_{2}, \ldots, \hat{\mathbf{f}}_{M}] \in \mathbb{R}^{M \times 8}$ contains the features of the selected segments.  
    Each $\hat{\mathbf{f}}_{i}$ is extracted from $\mathbf{F}$ according to the indication of $\mathbf{S}_b$, \textit{i.e.}, when the $i$-th element of $S_{b}$ is equal to 1.  
    \item {\textit{Binary Selection Mask}:} 
    Binary vector $\mathbf{S}_b \in \mathbb{R}^{N\times 1}$ indicates the current selection state of segments.
\end{itemize}

\textbf{Action:} Two discrete actions are defined as follows:1 ) \textit{Moving left} (``$\leftarrow$'', action ``0''); and 2) \textit{Moving right} (``$\rightarrow$'', action ``1''). Concretely, RSS-Net produces an action-probability matrix $\mathbf{A} \in \mathbb{R}^{M \times 2}$, where $A_{i,j} \in [0,1]$ denotes the probability that the $i$-th selected interpolation point chooses the $j$-th action. To mitigate the randomness bias of a single decision, each sample undergoes $B$ action iterations.  

\textbf{Reward Function:} Three reward functions are proposed and compared in this paper as follows:

{\textit{1.Benchmark-Calibrated Reward }(BCR):} 
    The error from the SOTA LSTM method~\cite{Ref27_Pang2019} is used as a benchmark as follows:
    \begin{equation}\label{eqt:brc}
    r = \frac{1}{N}\sum\limits_{i = 1}^{N} {r_i},
    \quad
    r_i =
    \begin{cases}
    1, & \text{if } \text{error}(t_i) < \text{error}(\hat{t}_i), \\
    0, & \text{otherwise},
    \end{cases}
    \end{equation}
    where $\hat{t}_i$ and $t_i$ are the result of LSTM and LRM in Eq.~\eqref{eqt:LR}, respectively, and function $\text{error}(t)$ measures the error of the predicted time $t$ by compared with the ground truth. Therefore, $r_i$ measures whether the ability of LRM is greater than that of LSTM or not. Eq.~\eqref{eqt:brc} encourages LRM has a better performance than LSTM, but has danger to barely obtain reward. 
    
\textit{2.Inverse Error Reward} (IER): The inverse error of selected by the RSS-Net serves as the reward, encouraging low prediction errors as follows:
    \begin{equation}
    r = \frac{1}{N}\sum\limits_{i = 1}^{N} \frac{1}{\text{error}(t_i)}.
    \label{eq:reward2}
    \end{equation}

\textit{3.Arrival Time difference Reward} (ATR):
ATP is the mean difference between the predicted arrival time of LRM and the ground-truth as follows:
    \begin{equation}
    r = \frac{1}{N}\sum\limits_{i = 1}^{N} |\text{error}(t_i)|.
    \label{eq:8}
    \end{equation}

\textbf{Avoid Selection Conflict:} 
To preserve the ordering of the selected segments and prevent index overlap during movement, positional constraints are imposed on selected segments. The upper bound index $\hat{I}_i$ for the $i$-th segment is defined as:
\begin{equation}
    \hat{I}_i =
    \begin{cases}
    \left\lceil \frac{I_i + I_{i+1}}{2} \right\rceil, & 1 \le i \le N-1, \\
    f, & i = N,
    \end{cases}
\label{4}
\end{equation}
where $\lceil \cdot \rceil$ denotes the ceiling operator, and $I_i$ is the index of the $i$-th selected segment.

Similarly, the lower bound index $\tilde{I}_i$ is defined as:
\begin{equation}
\tilde{I}_i =
\begin{cases}
\left\lceil \frac{I_{i-1} + I_i}{2} \right\rceil, & 2 \le i \le N, \\
0, & i = 1.
\end{cases}
\label{eq:5}
\end{equation}

Together, $\tilde{I}_i$ and $\hat{I}_i$ specify the allowable adjustment range for the $i$-th interpolation point as $[\tilde{I}_i, \hat{I}_i]$.  
After an action is executed, the index of the interpolation point is updated according to:
\begin{equation}
I_i' = I_i + a_i, \quad
a_i =
\begin{cases}
-\min(1,\, I_i - \tilde{I}_i), & \text{if action 0}, \\
\min(1,\, \hat{I}_i - I_i - 1), & \text{if action 1},
\end{cases}
\label{eq:action_conflict}
\end{equation}
where $a_i$ denotes the action-induced index shift.  
These constraints ensure that: the selected interpolation points remain properly ordered and non-overlapping.

\begin{small}\label{alg:RL}
\begin{algorithm}[t!]
    \caption{Deep Progressive Reinforcement Learning for Bus ETA Prediction}
    \label{alg:DPRL}
    \renewcommand{\algorithmicrequire}{\textbf{Input:}}
    \renewcommand{\algorithmicensure}{\textbf{Output:}}
    \begin{algorithmic}[1]
        \Require 
        Historical GPS trajectory data $\mathcal{D}$, 
        bus stop sequence $\mathcal{S}$, 
        interpolation point set $\mathcal{F}$, training epoch $E$.
        \Ensure 
        Trained model RSS-Net and LRM in~\eqref{eqt:LR}.
         \State Construct state $\mathbf{S}^0 = [\mathbf{S}^0_a, \mathbf{S}^0_b]$;
        \For{ $e = 0$ to $E$}
            
            \For{$j=1$ to $B$}
            \State RSS-Net output action probabilities $\mathbf{A}^j$, and avoid action conflicts by~\eqref{eq:action_conflict};
            \State Update $\mathbf{S}^{j-1}_b$ to $\mathbf{S}^j_b$ based on action $\mathbf{A}^j$;
            \State Update state $\mathbf{S}^e \leftarrow [{\mathbf{S}^j_a}, \mathbf{S}^j_b]$;
            \State LRM predicts arrival time $\hat{\mathbf{t}}$ at $j$-th inner iteration;
            \State Calculate the reward $r$;
            \EndFor
            \State Update state $\mathbf{S}^e \leftarrow [{\mathbf{S}_a}, \mathbf{S}_b]$;
            \State RSS-Net predicts action $\mathbf{A}$, and avoid action conflicts by Eq.~\eqref{eq:action_conflict};
            \State Update reward $r$;
            \State Update LRM in Eq.~\eqref{eqt:LR};
        \EndFor
        \State \Return Trained model RSS-Net and LRM in~\eqref{eqt:LR}.
    \end{algorithmic}
\end{algorithm}
\end{small}
 
\subsection{Loss Function and Optimization}

The network parameters are optimized using a cross-entropy based loss function as follows:
\begin{equation}
L({\bm{\theta}}) = -\frac{1}{B}\frac{1}{M} 
\sum_{j=1}^{B} \sum_{i=1}^{M} 
\log\big(\pi_{\bm{\theta}}(\mathbf{S}^j_i)\big) A^j_i r^j_i,
\label{eq:3-9}
\end{equation}
where $\pi_{\bm{\theta}}(\cdot)$ is the softmax function, which maps the network  with the parameter $\bm{\theta}$  into a probability distribution over actions.  
$\pi_{\bm{\theta}}(\mathbf{S}^j_i)$ denotes the probability of the action taken at the $i$-th point during the $j$-th iteration, $A^j_i$ is the executed action, and $r^j_i$ is the corresponding reward.  
$B$ denotes the number of action iterations performed for each training sample. Eq.~\eqref{eq:3-9} is optimized by policy-gradient computed from the accumulated rewards. Alg.~\ref{alg:RL} outlines the proposed method.  

\section{Data Analysis and Experimental Results} \label{sec:dataset_analysis}

\subsection{Dataset and Pre-processing}

\textbf{Dataset.} We use the dataset from~\cite{Ref27_Pang2019}, which consists of 23 consecutive days of bus operation data collected in February 2015. The first 22 days are used for training, while the remaining day serves as the testing set. In this study, we select 6 bus routes (No. 14, No. 60, No. 74, No. 110, No. 421, No. 563) in Beijing, each with varying operational lengths, providing a robust empirical foundation for validating the proposed model.

\textbf{ETA Interpolation.} For each trip, the distance between consecutive GPS points is first calculated, and the GPS trajectory is then transformed into distance–time data pairs. To estimate the ETA for each road segment, Kriging interpolation~\cite{bae2018missing}, a generalized form of Gaussian Processes, is used to interpolate the distance–time pairs $(d_i, t_i)$ for the $i$-th segment. In this study, each road segment is defined with a 100-meter interval.

\subsection{Experiment Settings}

In our experiments, we use the open-source deep learning framework PyTorch 1.8.0 with CUDA 11.1. All experiments are conducted on an Intel(R) i7-7700 CPU @ 3.60GHz with 32 GB of RAM and an NVIDIA 3080 GPU. During training, we train the models from scratch, without using any pre-trained weights. The batch size is set to 16, and we use Stochastic Gradient Descent (SGD) with a weight decay of 0.0005 and momentum of 0.9. The initial learning rate is set to $1 \times 10^{-3}$, which is decreased by a factor of 10 every 20 epochs after the first 40 epochs.

The initial segments are selected randomly, and the number of action iterations is set to 2 to ensure sufficient variation in the selected points. During testing, these selected segments, along with the bus stop locations, are arranged according to their spatial order and fed into the LRM for ETA prediction.

\subsection{Evaluation Metric}\label{Evaluation Metric}

The Mean Absolute Error (MAE) is used as the evaluation metric. MAE quantifies the average absolute deviation between the predicted and actual arrival times across all routes and operations, and is defined as:

\begin{equation}
MAE = \frac{2}{NV(V-1)} 
\sum_{i=1}^{N} 
\sum_{v=1}^{V-1} 
\sum_{\Delta=1}^{N-v} 
\left| t_{v+\Delta}^{i} - \hat{t}_{v+\Delta}^{i} \right|,
\label{eq:mae}
\end{equation}
where $i$ denotes the $i$-th trip, $N$ represents the total number of trips.

\subsection{Comparison Methods}

To comprehensively evaluate the effectiveness of the proposed method, several representative ETA prediction methods are compared:

\begin{itemize}
    \item K-Nearest Neighbor (KNN)~\cite{Ref18_Coffey2011} and Kernel Regression (KR)~\cite{Ref51x}:  
    Both methods are search-based approaches that rely on historical trajectory data. KNN selects the $k$ most similar trajectories to estimate the arrival time at the next stop, while KR computes nonlinear kernel-based weights to predict the ETA.
    
    \item Support Vector Machine on Probe Bus (SPB)~\cite{Ref54}:  
    This method uses probe vehicle data on downstream segments to model dynamic traffic characteristics using SVM-based regression.

    \item Linear Regression Model (LRM)~\cite{Ref51x}:  
    Assumes that bus travel times follow a multivariate Gaussian distribution. The predicted ETA is the posterior mean of the arrival time at the target stop.

    \item Additive Mixed Model (AMM)~\cite{Ref55}:  
    Considers multiple influencing factors and models the relationship between features and arrival times using nonlinear additive functions. Implemented using the \texttt{mgcv} package in R.

    \item Kalman Filter with Probe Bus (KFP)~\cite{Ref56}:  
    Utilizes GPS-based observations to estimate the ETA. The covariance factor in the filter is updated periodically using an adaptive algorithm.

    \item Multilayer Perceptron (MLP)~\cite{Ref57_Gurmu2014}:  
    A feedforward neural network model that uses the current location, timestamp, and distance to the target stop as inputs. Implemented using the \texttt{newff} function in MATLAB.

    \item Long Short-Term Memory (LSTM)~\cite{Ref27_Pang2019}:  
    A temporal deep learning model that captures sequential dependencies in bus trajectories. Implemented using the PyTorch framework with the same feature set as our proposed model.
\end{itemize}

\subsection{Empirical Results}

To verify the performance of the proposed model, bus routes are categorized based on their journey distances:  
1) Short-distance routes (0–10 km), including Bus 14 and Bus 60;  
2) Medium-distance routes (10–15 km), including Bus 110 and Bus 421;  
3) Long-distance routes ($> 15$ km), including Bus 74 and Bus 563.

All reported results correspond to the optimal configuration. The number of selected points in our method is set to 2/3 of all segments.  
Table~\ref{tab:mae_results} shows that the proposed method achieves the lowest MAE across all bus routes. The KFP method performs the worst due to its reliance on recursive corrections; when prior estimates deviate from the actual state, errors accumulate rapidly.  
The KNN and KR methods also exhibit limited accuracy, as they rely on the similarity between historical and current traffic conditions. Interestingly, the non-linear methods, i.e., LSTM and MLP, yield higher MAE errors than our method. For instance, our method outperforms the LSTM model by 0.69 for Bus Route 60, 0.54 for Bus Route 110, and 0.33 for Bus Route 563, respectively. 

Additionally, the following interesting observations can be drawn from the comparison:

\begin{itemize}
    \item \textit{Our method yields more significant improvements for short-distance journeys, but the improvement becomes less significant for long-distance journeys.} It is reasonable to assume that the non-linearity of long journeys is stronger than that of short ones. Therefore, the improvement achieved by our method is constrained by the linear prediction model in Eq.~\eqref{eqt:LR}.
    \item \textit{Our method outperforms non-linear methods.} The only explanation is that each road segment plays a different role in ETA prediction. Not every road segment is necessary or useful for ETA prediction.
\end{itemize}

\begin{table}[t]
\centering
\caption{Performance comparison of different methods. The best results are highlighted in bold, and the second-best results are underlined. The metric is MAE (min).}
\label{tab:mae_results}
\setlength{\tabcolsep}{1mm}{
\begin{tabular}{lcccccc}
\hline
\multirow{2}{*}{Methods} & \multicolumn{2}{c}{{Short distance }} & \multicolumn{2}{c}{{Mid distance }} & \multicolumn{2}{c}{{Long distance }} \\
\cline{2-7}
 & {14} & {60} & {110 } & {421} & {74} & {563} \\
\hline
LRM~\cite{Ref51x} & 1.58 & 1.58 & 1.72 & 1.30 & 1.52 & 1.26 \\
KNN~\cite{Ref18_Coffey2011} & 1.59 & 1.80 & 1.97 & 1.46 & 1.62 & 1.33 \\
KR~\cite{Ref51x} & 1.49 & 1.67 & 1.81 & 1.39 & 1.50 & 1.27 \\
AMM~\cite{Ref55} & 1.48 & 1.84 & 1.94 & 1.32 & 1.58 & 1.30 \\
SPB~\cite{Ref54} & \underline{1.43} & 1.85 & 2.03 & 1.37 & 1.66 & 2.92 \\
KFP~\cite{Ref56} & 3.22 & 4.15 & 4.87 & 2.72 & 3.44 & 3.28 \\
MLP~\cite{Ref57_Gurmu2014} & 1.57 & 1.55 & 2.15 & 3.63 & 1.54 & 1.94 \\
LSTM~\cite{Ref27_Pang2019} & 1.55 & \underline{1.50} & \underline{1.70} & \underline{1.28} & \underline{1.38} & \underline{1.19} \\
\hline
Our & \textbf{0.99} & \textbf{1.11} & \textbf{1.23} & \textbf{1.15} & \textbf{1.26} & \textbf{0.93} \\

\hline
\end{tabular}}
\end{table}

\subsection{Ablation Study on Reinforcement Learning}

To further validate the effectiveness of RSS-Net, an ablation experiment is conducted, comparing Random Selection (RS), RL-based selection, and the use of all interpolation points (ALL), where both RS and RL-based selection use the same proportion of interpolation points (2/3).  
Table~\ref{tab:ablation} shows that RS results in the poorest prediction accuracy among the three strategies. For instance, our method outperforms the RS method by 1.14 on route 60, 0.87 on route 14, and 0.83 on route 563.  
In contrast, the reinforcement learning network adaptively updates its policy based on predefined rewards, enabling it to consistently select the most representative interpolation points. Our method outperforms the ALL method by 0.59 on route 14, 0.49 on route 110, and 0.47 on route 60. As a result, the model achieves more accurate and stable ETA predictions across all bus routes.  
In summary, Table~\ref{tab:ablation} indicates that only a small number of special road segments significantly impact the ETA prediction for buses.

\begin{table}[t]
\centering
\caption{Performance comparison of different segment selection strategies (ALL, RS, and RL) across 6 bus routes.}
\label{tab:ablation}
\setlength{\tabcolsep}{1mm}{
\begin{tabular}{lcccccc}
\hline
\multirow{2}{*}{Methods} & \multicolumn{2}{c}{{Short distance }} & \multicolumn{2}{c}{{Mid distance }} & \multicolumn{2}{c}{{Long distance }} \\
\cline{2-7}
 & {14} & {60} & {110 } & {421} & {74} & {563} \\
\hline
ALL &\underline{1.58} & \underline{1.58} & \underline{1.72} & \underline{1.30} & \underline{1.52} & \underline{1.26} \\
RS       & 1.86 & 2.25 & 2.01 & 1.80 & 1.82 & 1.76 \\
RL & \textbf{0.99} & \textbf{1.11} & \textbf{1.23} & \textbf{1.15} & \textbf{1.26} & \textbf{0.93} \\
\hline
\end{tabular}}
\end{table}

\begin{figure}[h]
    \centering
    \includegraphics[width=0.45\textwidth]{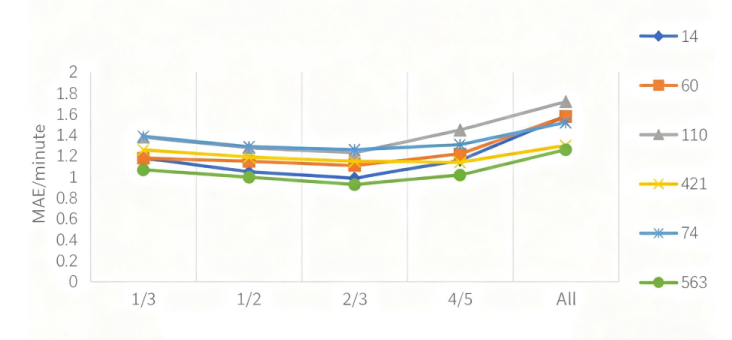}
    \caption{Comparison of MAE performance under different proportions of selected interpolation points.}
    \label{fig:interpolation_mae}
\end{figure}

This study investigates the impact of the number of selected road nodes on bus ETA prediction performance. As illustrated in Fig.~\ref{fig:interpolation_mae}, optimal prediction performance is achieved when two-thirds of the points are selected.

Specifically, when all interpolation points are utilized, the prediction error increases significantly compared to the scenario where only a few interpolation points are selected. Interestingly, when the number of interpolation points is reduced to one-third, the prediction error actually decreases. This comparison reveals that temporal correlation is less critical than the selection of road nodes.

The results suggest two key insights for bus ETA prediction:

\begin{itemize}
    \item \textit{Some segments are unnecessary for bus ETA prediction.} Although LRM in Eq.~\eqref{eqt:LR} is not an auto-regressive prediction method like LSTM, the comparison shows that some road segments can even have an adverse effect on the bus ETA prediction task.
    \item \textit{Diverse bus routes clearly highlight the necessity of RSS-Net in bus ETA prediction.} RSS-Net yields consistent performance improvements across diverse bus routes, indicating that our method exhibits robust generalization capabilities.
\end{itemize}

\begin{table}[t]
\centering
\caption{Comparisons among three reward strategies on bus routes No.110 and No.563 in terms of MAE (min).}
\label{tab:reward_design}
\begin{tabular}{lcc}
\toprule
{Strategy} & {110 } & {563 } \\
\midrule
BCR & 1.79 & 1.19 \\
IER & \underline{1.67} & \underline{1.12} \\
ATR & \textbf{1.23} & \textbf{0.93} \\
\bottomrule
\end{tabular}
\end{table}

\subsection{Experiment on Reward Selection Strategy}

Table~\ref{tab:reward_design} compares the effectiveness of three reward strategies for bus ETA prediction. Specifically, the ATR strategy achieves optimal performance, yielding the smallest MAE values of 1.23 for Route 110 and 0.93 for Route 563.
The underlying reason is that ATR can capture sufficient reward signals from all interpolation points, in contrast to the sparse reward signals provided by the BCR strategy. Notably, BCR is ill-suited for the bus ETA prediction task, as it only provides non-zero reward signals when the LRM outperforms its counterpart, the LSTM model. This scenario is unlikely when the LRM is not well-trained, whereas the ATR strategy enables RSS-Net to be effectively trained over time.

IMR generates reward values near unity, making the interpolation selection network insensitive to reward variation and prone to perceiving suboptimal selections as optimal. In contrast, ATR produces reward values with larger magnitude differences, allowing the network to better capture the importance of selected features.

\subsection{Effectiveness of Binary Mask}

Binary masking enables the network to identify which interpolation points have been selected. Table~\ref{tab:3-7} compares the MAE results of RSS-Net with and without the binary index mask. It shows that the binary mask significantly improves the accuracy of bus ETA prediction. Without the mask, the results are similar to those obtained through random selection. For instance, on route 110, the MAE for RSS-Net without the binary mask is 1.98, while the result for random selection is 2.01.

This comparison indicates that without the binary mask, the RL model fails to capture the distinguishing features. The introduction of the mask allows the network to recognize whether each interpolation point is selected, thereby learning positional dependencies and enhancing interpretability.

\begin{table}[h]
\centering
\caption{Comparisons RSS-Net with and without binary mask on bus routes No.110 and No.563 in terms of MAE(min.).}
\label{tab:3-7}
\begin{tabular}{ccc}
\hline
Methods & 110 & 563 \\ \hline
Random Selection & 2.01 & 1.76 \\\hline
RSS-Net w/o $\mathbf{S}_b$ & \underline{1.98} & \underline{1.11} \\
RSS-Net w $\mathbf{S}_b$ & \textbf{1.23} & \textbf{0.93} \\\hline
\end{tabular}
\end{table}

\subsection{Sensitivity of The number of Action Iterations }

To further evaluate the impact of action iterations on model performance, we tested the number of action iterations set to 2, 4, 6, and 8, as shown in Table~\ref{tab:iterations}. When the number of action iterations is set to 2, the prediction accuracy is highest, as compared in Table~\ref{tab:iterations}. This is because the road segments are sampled approximately every 100 meters, and a smaller number of action iterations allows for more frequent updates of interpolation information, thereby enabling better spatial feature correlation. Conversely, a larger number of action iterations leads to more abrupt changes in interpolation points, potentially missing optimal spatial feature distributions.

\begin{table}[t]
\centering
\caption{Comparisons  among different number of action iterations in terms of MAE(min).}
\label{tab:iterations}
\setlength{\tabcolsep}{1mm}{
\begin{tabular}{lcccccc}
\hline
\multirow{2}{*}{Methods} & \multicolumn{2}{c}{{Short distance }} & \multicolumn{2}{c}{{Mid distance }} & \multicolumn{2}{c}{{Long distance }} \\
\cline{2-7}
 & {14} & {60} & {110 } & {421} & {74} & {563} \\
\hline
2   & \textbf{0.99} & \textbf{1.11} & \textbf{1.23} & \textbf{1.15} & \textbf{1.26} & \textbf{0.93} \\
4   & \underline{1.01} & \underline{1.13} & \underline{1.25} & \underline{1.17} & \underline{1.27} & \underline{0.95} \\
6   & 1.04 & 1.14 & 1.26 & 1.19 & 1.30 & 0.96 \\
8   & 1.06 & 1.16 & 1.29 & 1.21 & 1.32 & 0.99 \\
\hline
\end{tabular}}
\end{table}

\begin{figure}[h]
    \centering
    \includegraphics[width=0.45\textwidth]{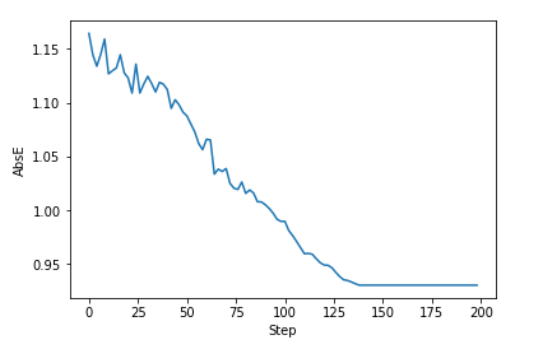}
    \caption{Convergence curve of MAE during the training process on Bus Route 563.}
    \label{fig:rl_mae_trend}
\end{figure}

\subsection{Training Efficiency}

We conduct a case study on Bus Route 563 (Xiangshan Park to Xizhimen) to validate the effectiveness of RSS-Net. As a long-distance route traversing diverse functional zones—including the Zhongguancun business district and Summer Palace tourist area—it exhibits high spatial heterogeneity and travel time variability . In our experiment, the model is initialized with randomly distributed interpolation points that fail to capture critical traffic patterns. To address this, RSS-Net employs a progressive refinement strategy, iteratively adjusting the segment positions by moving 2 steps per iteration driven by feedback rewards. Fig.~\ref{fig:rl_mae_trend} illustrates this optimization process: as the number of iterations increases, the agent successfully aligns with characteristic interpolation points, resulting in a significant and steady convergence of the prediction error (MAE).

\section{Conclusions} \label{sec:conclusion}

This paper presents a novel  DPRL framework for efficient bus ETA prediction. Unlike conventional methods that rely on uniform interpolation of GPS trajectories or computationally intensive deep neural networks, our model utilizes a lightweight linear predictor coupled with a  RSS-Net to adaptively identify informative non-uniform road segments . To address the issue of error accumulation in auto-regressive tasks, the proposed framework jointly models the segment selection as a sequential decision-making process, employing a composite reward function to enforce a balance between prediction accuracy and feature sparsity. Extensive experiments on large-scale real-world datasets demonstrate superior performance, validating that our "Less is More" strategy enables a simple linear model to outperform state-of-the-art non-linear baselines while significantly reducing feature dimensions. Statistical analysis confirms that our method exhibits robust generalization capabilities across bus routes of varying lengths and spatial complexities.

The demonstrated efficacy of our method in addressing the bus ETA prediction problem motivates its extension to other applications within digital transportation platforms, \textit{e.g.}, the highway ETA problem. Future work will focus on incorporating real-time contextual factors, \textit{e.g.}, traffic flow and weather, into a test-time adaptation~\cite{kim2024test} framework with RSS-Net for dynamically selecting pivot segments. Additionally, our results encourage the "prediction as generation" approach~\cite{rong2025generative} to focus on network modeling.

\bibliographystyle{IEEEtran}        

\bibliography{myref}

@ARTICLE{Ref3_chen2013,
  author = {Qin Chen and Elodie Adida and Jane Lin},
  title = {Implementation of an Iterative Headway-based Bus Holding Strategy with Real-time Information},
  journal = {Public Transport},
  year = {2013},
  volume = {4},
  number = {3},
  pages = {165--186}
}

@inproceedings{Ref18_Coffey2011,
  title={Time of Arrival Predictability Horizons for Public Bus Route},
  author={Coffey, Cathal and Pozdnoukhov, Alexei and Calabrese, Francesco},
  booktitle={ACM Sigspatial International Workshop on Computational Transportation Science, ACM},
  pages={1--5},
  year={2011}
}

@article{Ref27_Pang2019,
  title={Learning to Predict Bus Arrival Time From Heterogeneous Measurements via Recurrent Neural Network},
  author={Pang, Junbiao and Huang, Jing and Du, Yong and Yu, Haitao and Huang, Qingming and Yu, Baocai},
  journal={Intelligent Transportation Systems},
  volume={20},
  issue={9},
  pages={3283--3293},
  year={2019}
}

@inproceedings{Ref51x,
  author={Sinn, Mathieu and Yoon, Ji Won and Calabrese, Francesco and Bouillet, Eric},
  booktitle={2012 15th International IEEE Conference on Intelligent Transportation Systems}, 
  title={Predicting arrival times of buses using real-time GPS measurements}, 
  year={2012},
  volume={},
  number={},
  pages={1227-1232},
  doi={10.1109/ITSC.2012.6338767}}

@article{Ref54,
  title={Bus Arrival Time Prediction at Bus Stop with Multiple Routes},
  author={Yu, Bin and Lam, William and Tam, Mei Lam},
  journal={Transportation Research Part C Emerging Technologies},
  volume={19},
  issue={6},
  pages={1157--1170},
  year={2011}
}

@inproceedings{Ref55,
  title={Bus Travel Time Predictions Using Additive Models},
  author={Kormaksson, Matthias and Barbosa, Luciano and et~al.},
  booktitle={Data Mining},
  pages={875--880},
  year={2014}
}

@article{Ref56,
  title={Travel Time Prediction under Heterogeneous Traffic Conditions Using Global Positioning System Data from Buses},
  author={Devi, Vanajakshi Lelitha and Shankar, Subramanian and et~al.},
  journal={Intelligent Transport Systems},
  volume={3},
  issue={1},
  pages={1--9},
  year={2009}
}

@article{Ref57_Gurmu2014,
  title={Artificial Neural Network Travel Time Prediction Model for Buses Using Only GPS Data},
  author={Gurmu, Zegeye Kebede and Fan, Wei},
  journal={Journal of Public Transportation},
  volume={17},
  issue={2},
  pages={45--65},
  year={2014}
}

@ARTICLE{Ref60,
  author={Tang, Yihong and Zhao, Zhan and Deng, Weipeng and Lei, Shuyu and Liang, Yuebing and Ma, Zhenliang},
  journal={IEEE Transactions on Intelligent Transportation Systems}, 
  title={RouteKG: A Knowledge Graph-Based Framework for Route Prediction on Road Networks}, 
  year={2025},
  volume={},
  number={},
  pages={1-19},
  doi={10.1109/TITS.2025.3615448}}

@ARTICLE{Ref70,
  author={Liu, Tao and Jiang, Aimin and Miao, Xiaoyu and Tang, Yibin and Zhu, Yanping and Kwan, Hon Keung},
  journal={IEEE Sensors Journal}, 
  title={Graph-Based Dynamic Modeling and Traffic Prediction of Urban Road Network}, 
  year={2021},
  volume={21},
  number={24},
  pages={28118-28130},
  doi={10.1109/JSEN.2021.3124818}}

@ARTICLE{Ref72,
  author = {Zhang, X. and Yan, M. and Xie, B. and Yang, H. and Ma, H.},
  journal = {Advances in Engineering Information},
  title = {An automatic real-time bus schedule redesign method based on bus arrival time prediction},
  year = {2021},
  volume = {48},
  pages = {101295},
  doi = {10.1016/j.aei.2021.101295},
}

@ARTICLE{Ref73,
  author = {Mishalani, R. G. and McCord, M. M. and Wirtz, J.},
  journal = {Journal of Public Transportation},
  title = {Passenger Wait Time Perceptions at Bus Stops: Empirical Results and Impact on Evaluating Real-Time Bus Arrival Information},
  year = {2006},
  volume = {9},
  number = {2},
  pages = {89--106},
  doi = {10.5038/2375-0901.9.2.5},
}

@INPROCEEDINGS{Ref74,
  author = {Maiti, S. and Pal, A. and Chattopadhyay, T. and Mukherjee, A.},
  title = {Historical data-based real-time prediction of vehicle arrival time},
  booktitle = {Proceedings of the 17th IEEE International Conference on Intelligent Transportation Systems (ITSC)},
  year = {2014},
  pages = {1837--1842},
  doi = {10.1109/ITSC.2014.6957960},
}

@ARTICLE{Ref82,
  author = {Roy, T. S. and Roy, J. K. and Mandal, N.},
  journal = {Biomedical Engineering Advances},
  title = {Deep Encoder Cross Network for ETA prediction},
  year = {2022},
  volume = {3},
  pages = {100035}
}

@ARTICLE{Ref83,
  author = {Ahsan, M. M. and Siddique, Z.},
  journal = {Artificial Intelligence in Medicine},
  title = {FMA-ETA: Feedforward attention model for ETA prediction},
  year = {2022},
  volume = {128},
  pages = {102289}
}

@INPROCEEDINGS{Ref86,
  author = {Jiho Park and Tong Liu and Chieh Wang and Hong Wang and Qichao Wang and Zhong-Ping Jiang},
  title = {Traffic Signal Control for Large-Scale Urban Traffic Networks: Real-World Experiments using Vision-based Sensors},
  booktitle = {2024 IEEE 18th International Conference on Control \& Automation (ICCA)},
  year = {2024},
  pages = {282--287}
}

@ARTICLE{Ref87,
  author = {Amit Chougule and Vinay Chamola and Vikas Hassija and Pranav Gupta and Fei Richard Yu},
  title = {A Novel Framework for Traffic Congestion Management at Intersections Using Federated Learning and Vertical Partitioning},
  journal = {IEEE Transactions on Consumer Electronics},
  year = {2024},
  volume = {70},
  number = {1},
  pages = {1725--1735}
}

@INPROCEEDINGS{Ref88,
  author = {Pengfei Du and Liang Dai and Yingying Ding and Hongke Xu},
  title = {Performance Analysis of Deep Reinforcement Learning-based Traffic Signal Control System at Urban Bottleneck Sections},
  booktitle = {2023 9th International Conference on Mechanical and Electronics Engineering (ICMEE)},
  year = {2023},
  pages = {247--252}
}

@INPROCEEDINGS{Ref90,
  author = {Yining Ma and Qadeer Khan and Daniel Cremers},
  title = {Multi Agent Navigation in Unconstrained Environments using a Centralized Attention based Graphical Neural Network Controller},
  booktitle = {2023 IEEE 26th International Conference on Intelligent Transportation Systems (ITSC)},
  year = {2023},
  pages = {2893--2900}
}

@INPROCEEDINGS{Ref92,
  author={Wang, Peng and Wei, Xiang and Hu, Fangxu and Han, Wenjuan},
  booktitle={2024 International Conference on Computational Linguistics and Natural Language Processing (CLNLP)}, 
  title={TransGPT: Multi-modal Generative Pre-trained Transformer for Transportation}, 
  year={2024},
  volume={},
  number={},
  pages={96-100},
  doi={10.1109/CLNLP64123.2024.00026}}

@article{VARGHESE20252618,
title = {Transportation infrastructure and economic growth: Does there exist causality and spillover? A Systematic Review and Research Agenda},
journal = {Transportation Research Procedia},
volume = {82},
pages = {2618-2632},
year = {2025},
issn = {2352-1465},
doi = {https://doi.org/10.1016/j.trpro.2024.12.208},
author = {Ann Mary Varghese and Rudra Prakash Pradhan},
}

@article{LU2025103461,
title = {Multi-driver transportation scheduling for improving supply chain resilience},
journal = {Omega},
pages = {103461},
year = {2025},
issn = {0305-0483},
doi = {https://doi.org/10.1016/j.omega.2025.103461},
author = {Shaojun Lu and Yiyu Song and Min Kong and Chaoming Hu and Amir M. Fathollahi-Fard and Maxim A. Dulebenets},
}

@article{XIAO2024103787,
title = {Sustainable and robust route planning scheme for smart city public transport based on multi-objective optimization: Digital twin model},
journal = {Sustainable Energy Technologies and Assessments},
volume = {65},
pages = {103787},
year = {2024},
issn = {2213-1388},
doi = {https://doi.org/10.1016/j.seta.2024.103787},
author = {Ming Xiao and Lihua Chen and Haoxiong Feng and Zhigao Peng and Qiong Long},
}

@article{SERIN2022111403,
title = {Predicting bus travel time using machine learning methods with three-layer architecture},
journal = {Measurement},
volume = {198},
pages = {111403},
year = {2022},
issn = {0263-2241},
doi = {https://doi.org/10.1016/j.measurement.2022.111403},
author = {Faruk Serin and Yigit Alisan and Metin Erturkler},
}

@article{SHEN2025127622,
title = {A novel model incorporating deep learning and Kalman filter augmentation for route-level bus arrival time prediction with error accumulation mitigation},
journal = {Expert Systems with Applications},
volume = {281},
pages = {127622},
year = {2025},
issn = {0957-4174},
doi = {https://doi.org/10.1016/j.eswa.2025.127622},
author = {Jinxing Shen and Qinxin Liu and Yining Zhang and Miao Yu},
}

@article{PETERSEN2019426,
title = {Multi-output bus travel time prediction with convolutional LSTM neural network},
journal = {Expert Systems with Applications},
volume = {120},
pages = {426-435},
year = {2019},
issn = {0957-4174},
doi = {https://doi.org/10.1016/j.eswa.2018.11.028},
author = {Niklas Christoffer Petersen and Filipe Rodrigues and Francisco Camara Pereira},
}

@article{MA2019536,
title = {Bus travel time prediction with real-time traffic information},
journal = {Transportation Research Part C: Emerging Technologies},
volume = {105},
pages = {536-549},
year = {2019},
issn = {0968-090X},
doi = {https://doi.org/10.1016/j.trc.2019.06.008},
author = {Jiaman Ma and Jeffrey Chan and Goce Ristanoski and Sutharshan Rajasegarar and Christopher Leckie},
}

@article{KHAIRY2025431,
title = {Adaptive traffic prediction model using Graph Neural Networks optimized by reinforcement learning},
journal = {International Journal of Cognitive Computing in Engineering},
volume = {6},
pages = {431-440},
year = {2025},
issn = {2666-3074},
doi = {https://doi.org/10.1016/j.ijcce.2025.02.001},
author = {Mohammed Khairy and Hoda M.O. Mokhtar and Mohammed Abdalla},
}

@article{WANG202171,
title = {Automatic generation of large-scale 3D road networks based on GIS data},
journal = {Computers \& Graphics},
volume = {96},
pages = {71-81},
year = {2021},
issn = {0097-8493},
doi = {https://doi.org/10.1016/j.cag.2021.02.004},
author = {Hua Wang and Yue Wu and Xu Han and Mingliang Xu and Weizhe Chen},
}

@article{khan2025deep,
  title={Deep learning model for efficient traffic forecasting in intelligent transportation systems},
  author={Khan, Shakir and Alghayadh, Faisal Yousef and Ahanger, Tariq Ahamed and Soni, Mukesh and Viriyasitavat, Wattana and Berdieva, Uguloy and Byeon, Haewon},
  journal={Neural Computing and Applications},
  volume={37},
  number={20},
  pages={14673--14686},
  year={2025},
  publisher={Springer}
}

@article{zhao2019t,
  title={T-GCN: A temporal graph convolutional network for traffic prediction},
  author={Zhao, Ling and Song, Yujiao and Zhang, Chao and Liu, Yu and Wang, Pu and Lin, Tao and Deng, Min and Li, Haifeng},
  journal={IEEE transactions on intelligent transportation systems},
  volume={21},
  number={9},
  pages={3848--3858},
  year={2019},
  publisher={IEEE}
}

@article{JI2020578,
title = {Reasoning Traffic Pattern Knowledge Graph in Predicting Real-Time Traffic Congestion Propagation},
journal = {IFAC-PapersOnLine},
volume = {53},
number = {5},
pages = {578-581},
year = {2020},
note = {3rd IFAC Workshop on Cyber-Physical \& Human Systems CPHS 2020},
issn = {2405-8963},
doi = {https://doi.org/10.1016/j.ifacol.2021.04.145},
author = {Qingyuan Ji and Junchen Jin},
}

@article{LAI2024100164,
title = {Reinforcement learning in transportation research: Frontiers and future directions},
journal = {Multimodal Transportation},
volume = {3},
number = {4},
pages = {100164},
year = {2024},
issn = {2772-5863},
doi = {https://doi.org/10.1016/j.multra.2024.100164},
author = {Xiongfei Lai and Zhenyu Yang and Jiaohong Xie and Yang Liu},
}

@article{farazi2021deep,
  title={Deep reinforcement learning in transportation research: A review},
  author={Farazi, Nahid Parvez and Zou, Bo and Ahamed, Tanvir and Barua, Limon},
  journal={Transportation research interdisciplinary perspectives},
  volume={11},
  pages={100425},
  year={2021},
  publisher={Elsevier}
}

@article{bae2018missing,
  title={Missing data imputation for traffic flow speed using spatio-temporal cokriging},
  author={Bae, Bumjoon and Kim, Hyun and Lim, Hyeonsup and Liu, Yuandong and Han, Lee D and Freeze, Phillip B},
  journal={Transportation Research Part C: Emerging Technologies},
  volume={88},
  pages={124--139},
  year={2018},
  publisher={Elsevier}
}

@article{kim2024test,
  title={Test-time adaptation induces stronger accuracy and agreement-on-the-line},
  author={Kim, Eungyeup and Sun, Mingjie and Baek, Christina and Raghunathan, Aditi and Kolter, J Zico},
  journal={Advances in Neural Information Processing Systems},
  volume={37},
  pages={120184--120220},
  year={2024}
}

@article{rong2025generative,
  title={Generative artificial intelligence in intelligent transportation systems: A systematic review of applications},
  author={Rong, Rui and Ma, Shoufeng and Ren, Nianlu and Lin, Qinping and Jia, Ning},
  journal={Frontiers of Engineering Management},
  pages={1--17},
  year={2025},
  publisher={Springer}
}

\end{document}